\documentclass[review,authoryear]{elsarticle}
\usepackage{graphicx}
\usepackage{times}
\usepackage{textcomp}
\usepackage{multirow}
\usepackage{bm}
\usepackage{booktabs}
\usepackage{amsthm}
\usepackage{amstext}
\usepackage{algorithmic,algorithm}
\usepackage{tikz}
\usepackage{pgfplots}
\usepackage{subfigure}
\usetikzlibrary{decorations.markings}
\usetikzlibrary{trees}
\usetikzlibrary{shapes}
\tikzstyle{nodo}=[ellipse,draw=black!60,fill=black!10,line width=.7pt,minimum width=1.1cm,minimum height=.6cm]
\tikzstyle{nodo2}=[ellipse,draw=black!60,fill=red!10,line width=.7pt,minimum width=1.1cm,minimum height=.6cm]
\tikzstyle{vuoto}=[]
\tikzstyle{arco}=[draw=black!80,line width=.7pt, postaction={decorate}, decoration={markings,mark=at position 1.0 with {\arrow[
draw=black!80,line width=.7pt]{>}}}]
\pgfplotsset{tick label style={font=\footnotesize},
label style={font=\footnotesize},
legend style={font=\footnotesize}}
\tikzset{every mark/.append style={scale=0.7}}
\usetikzlibrary{plotmarks}
\makeatletter
\usepackage{color,soul}
\sethlcolor{lightgray}
\setulcolor{red}
\makeatother
\usepackage[english]{babel}

\title{Credal Classification based on AODE and compression coefficients}
\author[idsia]{G. Corani\corref{cor1}}
\author[idsia]{A. Antonucci}
\address[idsia]{IDSIA\\
Istituto Dalle Molle di Studi sull'Intelligenza Artificiale\\
CH-6928 Manno (Lugano), Switzerland}
\cortext[cor1]{Corresponding author: giorgio@idsia.ch}

\begin{document}
% \title{Credal Classification based on AODE and compression coefficients}
% \author{Giorgio Corani\and Alessandro Antonucci}
% \institute{Giorgio Corani\at IDSIA (Istituto Dalle Molle di Studi sull'Intelligenza Artificiale), Manno - Lugano (Switzerland)\\ \email{giorgio@idsia.ch}           
%            \and Alessandro Antonucci\at IDSIA (Istituto Dalle Molle di Studi sull'Intelligenza Artificiale), Manno - Lugano (Switzerland)\\ \email{alessandro@idsia.ch}}
% \maketitle
\begin{abstract}
Bayesian model averaging (BMA) is a common approach to average over alternative models; yet, it usually gets  excessively concentrated around the single most probable model, therefore achieving only sub-optimal classification performance. The compression-based approach \citep{boulle2007compression}  overcomes this problem; it averages over the different models by applying a logarithmic smoothing over the models' posterior probabilities. This approach has shown excellent performances when applied to ensembles of naive Bayes classifiers. AODE is another ensemble of models with high performance \citep{webb2005not}: it consists of a collection of non-naive classifiers (called SPODE) whose probabilistic predictions are aggregated by simple arithmetic mean.
Aggregating the SPODEs via BMA rather than by arithmetic mean deteriorates the performance; instead, we propose to aggregate the SPODEs via the compression coefficients and we show that the resulting classifier obtains a slight but consistent improvement over AODE. However, an important issue in any Bayesian ensemble of models is the arbitrariness in the choice of the prior over the models. We address this problem by adopting the paradigm of \emph{credal} classification, namely by substituting the unique prior with a set of priors.  Credal classifier are able to automatically recognize the \textit{prior-dependent} instances, namely the instances whose most probable class varies, when different priors are considered; in these cases, credal classifiers remain reliable by returning a set of classes rather than a single class. We thus develop the credal version of both the BMA-based and the compression-based ensemble of SPODEs, substituting the single prior over the models by a set of priors. By experiments we show that both credal classifiers provide overall higher classification reliability than their determinate counterparts. Moreover, the compression-based credal classifier compares favorably to previous credal classifiers.
\end{abstract}

\maketitle
\section{Introduction}\label{sec:intro}
Bayesian model averaging (BMA) \citep{hoeting1999bma} is a sound solution to the uncertainty which characterizes the identification of the
supposedly best model for a certain data set; given a set of alternative models, BMA weights the inferences produced by the various models,
using the models' posterior probabilities as weights.
BMA assumes the data to be generated by one of the considered models; under this assumption, it provides better predictive accuracy than any single model \citep{hoeting1999bma}. However, such an assumption is generally not true; for this reason, on real data sets BMA does not generally perform very well;
see the discussion and the references in  \cite{cerquides2005robust} for more details.
The problem is that BMA gets excessively concentrated around the single most probable model \citep{domingos2000bac,minkaBMA}: especially on large data sets, ``\textit{averaging using the posterior probabilities to weight the models is almost the same as selecting the MAP model}'' \citep{boulle2007compression}.
To overcome the problem of BMA getting excessively concentrated around the most probable model, a \textit{compression-based} approach has been introduced in \citep{boulle2007compression}; it computes more evenly-distributed weights, by applying a logarithmic smoothing to the models posterior probabilities. The compression-based weights, which can be justified from an information-theoretic viewpoint, have been used in \cite{boulle2007compression} to average over different naive Bayes classifiers, characterized by different feature sets, obtaining excellent rank in international competitions on classification.

Another ensemble of Bayesian networks classifiers known for its good performance is AODE \citep{webb2005not}, which is instead based on
a set  of SPODE (SuperParent-One-Dependence Estimator) models. Each SPODE adopts a certain feature as a \textit{super-parent}, namely it models all the remaining features as depending on both the class \textit{and} the super-parent. AODE then simply averages the posterior probabilities computed by the different SPODEs.
Alternative methods to aggregate SPODEs, more complex than AODE,  have been considered \citep{yang2007select}, but  AODE generally outperforms them: ``\textit{AODE, which simply linearly combines every SPODE without any selection or weighting, is actually more effective than the majority of rival schemes}''.
As reported in \citep{cerquides2005robust,yang2007select}, AODE outperforms aggregating SPODEs via BMA; in both \citep{yang2007select, cerquides2005robust} the best results were instead obtained using an algorithm (called MAPLMG), which estimates the most probable linear \textit{mixture} of  SPODEs; this overcomes the problem of assuming a single SPODE to be the true model. In this paper, we address this problem by means of the compression coefficients.

As a preliminary step we develop BMA-AODE, namely BMA over SPODEs, with some computational differences with respect to the framework of \cite{yang2007select} and \cite{cerquides2005robust}; our results confirm however that BMA over SPODEs is outperformed by AODE. Then we develop the novel COMP-AODE classifier, which weights the SPODEs using the compression-based coefficients, and we show that it yields a slight but consistent improvement in the classification performance over the standard AODE. Considering the high performance of AODE, we regard this result as noteworthy.

An important issue in any Bayesian ensemble is choosing the prior over the models. A common choice is to adopt a uniform mass function, as we do in both BMA-AODE and COMP-AODE; this however can be criticized from different standpoints; see for instance the rejoinder in  \cite{hoeting1999bma}. In \cite{boulle2007compression}, a prior which favors simpler models over complex ones is adopted. Although all these choices are reasonable, the specification of any single prior implies some arbitrariness, which entails the risk of prior-dependent, and hence potentially fragile, conclusions.

In fact, the specification of the prior over the models  is a serious open problem for Bayesian ensembles of models.
We address this problem by adopting  the paradigm of \textit{credal classification}  \citep{corani_chapter2012,corani2008a}, namely dropping the unique prior in favor of a \textit{set} of priors (prior \textit{credal set}) \citep{levi1980}. While  a traditional non-informative priors represents a condition of \textit{indifference} between the alternative models, a credal set  describes a condition of prior \textit{ignorance}, letting thus vary  the prior probability of each model over a wide interval, instead of fixing it  to a specific number.
Credal classifiers are able to automatically detect the instances whose most probable class varies when different priors are considered;
such instances are called \textit{prior-dependent}. Credal classifiers remain reliable on prior-dependent instances by returning a \textit{set} of classes;  traditional classifiers have  instead typically  low accuracy on the prior-dependent instances \citep{corani2008d,corani2008a}.

We then develop BMA-AODE* and COMP-AODE*, namely the credal counterparts of respectively BMA-AODE and COMP-AODE.  By extensive experiments we show that both credal classifiers  are sensible extension of their single-prior counterparts; in fact, they return a small-sized but highly accurate set of classes on the prior-dependent instances, over which instead their single-prior counterparts have reduced accuracy. We conclude by showing that  COMP-AODE* compares favorably to both BMA-AODE* and other existing credal classifiers.

\section{Methods}\label{sec:bg}
We consider a classification problem with $k$ \emph{features}; we denote by $C$ the \textit{class} variable (taking values in  $\mathcal{C}$) and by $\mathbf{A}:=(A_1,\ldots,A_k)$ the set of features, taking values respectively in $\mathcal{A}_1,\ldots,\mathcal{A}_k$.  For a generic variable $A$, we denote as $P(A)$ the probability mass function over its values and as $P(a)$ the probability that $A=a$. We assume the data to be complete and the training data $\mathcal{D}$ to contain $n$ instances. We learn the model parameters from the training data by adopting Dirichlet priors and setting the equivalent sample size to 1.
% When this step has been accomplished, the classifier has estimated from $\mathcal{D}$ a joint probability mass function $P(C,\mathbf{A})$.
 Under 0-1 loss a traditional probabilistic classifier returns, for a test instance $\bf{\tilde{a}}$ $=\{\tilde{a_1},\ldots,\tilde{a_k}\}$ whose class is unknown, the most probable class  $c^*$:
\[
c^*:=\arg\max_{c\in\mathcal{C}} P(c|\mathbf{\tilde{a}}).
\]
Classifiers based on imprecise-probabilities (\emph{credal} classifiers) change this paradigm, by occasionally returning more classes; this happens in particular when the most probable class is \textit{prior-dependent}. We discuss this point more in detail later, when presenting credal classifiers.

\subsection{From Naive Bayes to AODE}
\begin{figure}[ht!]
\centering
\subfigure[Naive Bayes.]{
\begin{tikzpicture}[scale=.7]\tiny
\node[nodo] (x0)  at (0,3) {$C$};
\node[nodo] (x1)  at (-3,0) {$A_1$};
\node[nodo] (x2)  at (-1,0) {$A_2$};
\node[nodo] (x3)  at (1,0) {$A_3$};
\node[nodo] (x4)  at (3,0) {$A_4$};
\draw[arco] (x0) -- (x1);
\draw[arco] (x0) -- (x2);
\draw[arco] (x0) -- (x3);
\draw[arco] (x0) -- (x4);
\end{tikzpicture}
\label{fig:subfigNBC}
}
\hspace{.5cm}
\subfigure[A possible TAN structure.]{
\begin{tikzpicture}[scale=.7]\tiny
\node[nodo] (x0)  at (0,3) {$C$};
\node[nodo] (x1)  at (-3,0) {$A_1$};
\node[nodo] (x2)  at (-1,0) {$A_2$};
\node[nodo] (x3)  at (1,0) {$A_3$};
\node[nodo] (x4)  at (3,0) {$A_4$};
\draw[arco] (x0) -- (x1);
\draw[arco] (x0) -- (x2);
\draw[arco] (x0) -- (x3);
\draw[arco] (x0) -- (x4);
\draw[arco] (x1) -- (x2);
\draw[arco] (x2) -- (x3);
\draw[arco] (x4) -- (x3);
\end{tikzpicture}
\label{fig:subfigTAN}
}
\caption{\label{fig:nbc}Naive Bayes vs TAN.}
\end{figure}
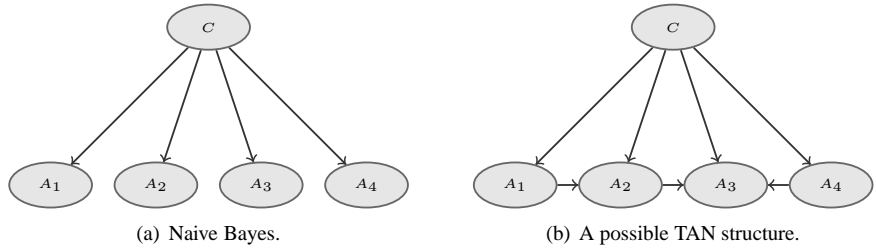

The Naive Bayes classifier  assumes the  stochastic independence of the features given the class; it therefore factorizes the joint probability as follows:
\begin{equation}\label{eq:nbcjoint}
P(c,\mathbf{a}):= P(c) \cdot \prod_{j=1}^k P(a_j|c),
\end{equation}
 corresponding to the topology of Fig.\ref{fig:subfigNBC}. Despite the biased estimate of probabilities due to the above (so-called \textit{naive}) assumption, naive Bayes performs well under 0-1 loss \citep{domingos1997}; it thus constitutes  a reasonable choice if the goal is simple classification, without the need for accurate probability estimates; it is especially competitive on data sets of small and medium size , thanks to its low variance error \citep{friedmanJ1997}.

To improve the model, weaker assumptions about the conditional independence of the features have to be considered; for instance, the tree-augmented naive classifier (TAN) allows each feature to depend on the class \textit{and} on possibly another single feature, constraining however the subgraph involving only  the features to be a tree; an example is shown in Fig.\ref{fig:subfigTAN}. Generally, TAN outperforms naive Bayes in classification \citep{friedmanN1997a}.

\begin{figure}[htp!]
\centering
\begin{tikzpicture}[scale=.7]\tiny
\node[nodo] (x0)  at (0,3) {$C$};
\node[nodo] (x1)  at (-3,0) {$A_1$};
\node[nodo] (x2)  at (-1,0) {$A_2$};
\node[nodo] (x3)  at (1,0) {$A_3$};
\node[nodo] (x4)  at (3,0) {$A_{4}$};
\draw[arco] (x0) -- (x1);
\draw[arco] (x0) -- (x2);
\draw[arco] (x0) -- (x3);
\draw[arco] (x0) -- (x4);
\draw[arco] (x1) -- (x2);
\draw[arco] (x1) .. controls (-1,-.7) .. (x3);
\draw[arco] (x1) .. controls (-1,-1) .. (x4);
\end{tikzpicture}
\caption{\label{fig:aode} SPODE with super-parent $A_1$.}
\end{figure}
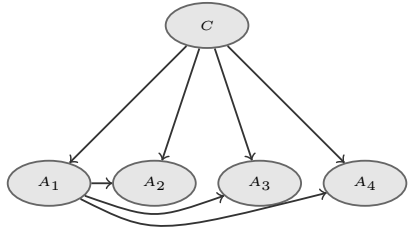

The AODE classifier \citep{webb2005not} is an ensemble of $k$ SPODE (SuperParent One Dependence Estimator) classifiers;
each SPODE is characterized by a certain \textit{super-parent } feature, so that the other features are modeled as depending on both the class and the super-parent, as shown in in Fig.\ref{fig:aode}. In fact, each single SPODE is a TAN.
\par
We denote the set of SPODEs as $\mathcal{S}:=\{ s_1,\ldots,s_k\}$, where  $s_j$ indicates the  SPODE with super-parent $A_j$.  The joint probability of SPODE $s_j$ factorizes as:
$$
P(c,\mathbf{a}|s_j) =  P(c) \cdot P(a_j|c) \cdot \prod_{l=1..k,l\neq j}^k P(a_l|a_j,c).
$$
In order to classify the test instance $\bf{\tilde{a}}$, AODE averages the posterior probability $P(c|\mathbf{\tilde{a}},s_j)$ computed by each single SPODE:
$$
P(c|\mathbf{a}) = \frac{1}{k}\sum_{j=1}^{j=k}P(c,\mathbf{a}|s_j)
$$
 In this paper we focus on more sophisticated approaches for aggregating the predictions of the SPODEs.

\subsection{Bayesian Model Averaging  (BMA) with SPODEs}
BMA assumes that one of the models in the ensemble is the true one. Under this assumption, the optimal strategy is to weight the inferences produced by the models of the ensemble using as weights the models' posterior probabilities. By applying BMA on top of different SPODEs, we thus assume one of the SPODEs to be the true model. We thus introduce a variable $S$ over $\mathcal{S}$, where $P(S=s_j)$ denotes the \textit{prior} probability of SPODE $s_j$ to be the true model. Considering that every SPODE has the same number of variables, the same number of arcs and the same in-degree\footnote{The in-degree is the maximum number of parents per node: it is two for any SPODE.}, we adopt a \textit{uniform} prior, thus assigning prior probability $1/k$ to each SPODE. In fact, the uniform prior over the models is frequently adopted within BMA. To classify the test instance $\bf{\tilde{a}}$, BMA computes the following posterior mass function:
$$P(c|\tilde{\mathbf{a}}) = \sum_{j=1}^k  P(c|\tilde{\mathbf{a}},s_j) \cdot P(s_j|\mathcal{D})
\propto
\sum_{j=1}^k
P(c|\tilde{\mathbf{a}},s_j) \cdot P(\mathcal{D}|s_j) \cdot P(s_j),
\label{eq:bma1}
$$
where $P(\mathcal{D}|s_j)$ is the \textit{marginal} likelihood of  $s_j$, namely
\[P(\mathcal{D}|s_j)=\int P(\mathcal{D}|s_j,
\theta_j) \cdot P(\theta_j|s_j) \cdot d \theta_j,\] with $\theta_j$ denoting the parameters of SPODE $s_j$.
This computational schema has been adopted to implement BMA over SPODEs in \citep{cerquides2005robust,yang2007select}, and has been outperformed by AODE.

 The marginal likelihood measures how good the model is at representing the \textit{joint} distribution; yet, a classifier has instead to estimate the posterior probability of the classes \textit{conditionally} on the features.  Therefore, a model can perform badly at classification despite having high marginal likelihood \citep{cowell2001searching,kontkanen1999supervised}; for this reason, scoring rules more appropriate for classification should be considered. Following \cite{boulle2007compression}, we thus substitute the marginal likelihood with \textit{conditional} likelihood:
\begin{equation}\label{eq:cond_lik}
L_j:=\prod_{i=1}^{n} P(c^{(i)}|\mathbf{a}^{(i)},s_j,\hat{\theta}_j),
\end{equation}
where $P(c^{(i)}|\mathbf{a}^{(i)},s_j,\hat{\theta}_j)$ denotes the probability assigned by model $s_j$ to the true class of the $i$-th instance, and $\hat{\theta}_j$ is the  estimate of the parameters of model $s_j$.
% Preliminary experiments showed in fact a slight improvement of performance due to the adoption of the conditional likelihood instead of the marginal one.

We call BMA-AODE the classifier which estimates the posterior probabilities of the class, given the test instance $\tilde{\mathbf{a}}$, as follows:
\begin{equation}\label{eq:bma}
P(c|\tilde{\mathbf{a}}) \propto \sum_{j=1}^k P(c|\tilde{\mathbf{a}},s_j) \cdot L_j \cdot P(s_j).
\end{equation}

Especially on large data sets, the difference between the likelihoods of the different SPODEs might be of several order of magnitudes. We remove from the ensemble  the SPODEs whose conditional likelihood is smaller than $L_{\mathrm{max}}/10^4$, where $L_{\mathrm{max}}$ is the maximum conditional likelihood among all SPODEs; discarding models with very low posterior probability is in fact common when dealing with BMA; this procedure can be seen as a  \emph{belief revision} \citep{dubois1997}. Given the joint beliefs $P(X,Y)$, the \textit{revision} $P'(X,Y)$ induced by a marginal $P'(Y)$ is defined by $P'(x,y):=P(x|y)\cdot P'(y)$. In other words, if $P'(y)$ is known to be a better model than $P(y)$ for the marginal beliefs about $y$, this information can be used in the above described way to redefine the joint. Accordingly, in BMA-AODE, the marginal beliefs about $S$ have been replaced by a better candidates, inducing a revision in the corresponding joint model.

\subsubsection{Exponentiation of the Log-Likelihoods}
Regardless whether the marginal likelihood or the conditional likelihood is considered, it is common to compute the  \textit{log-likelihood} rather than the likelihood, in order to avoid numerical problems due to the multiplication of many probabilities. However, if the log-likelihoods are very negative, as it happen on large data sets, their exponentiation can suffer numerical problems too. This issue has been addressed in \cite{yang2007select} by means of high numerical precision: ``\textit{BMA often lead to arithmetic overflow when calculating very large exponentials or factorials. One solution is to use the Java class BigDecimal which unfortunately can be very slow.}'' Algorithm \ref{algo-lliks} describes a procedure for exponentiating the log-likelihoods, which is both numerically robust and computationally fast.  The procedure has been communicated to us by D. Dash, who published several works on BMA \citep{dash2004jmlr}.

\begin{algorithm}[!ht]
\caption{Robust exponentiation of log-likelihoods.}
\label{algo-lliks}
\begin{algorithmic}
\STATE
\REQUIRE {Array \texttt{log\_liks} of log-likelihoods, assumed of length \texttt{k}.}
\STATE
\STATE \texttt{minVal=min(log\_liks)}
\STATE
\FOR{\texttt{i = 1:k}}
\STATE \texttt{shifted\_logliks(i)=logliks(i)-minVal;}\\
\STATE  \texttt{tmp\_liks(i)=exp(shifted\_logliks(i));}\\
\ENDFOR
\STATE
\STATE \texttt{total=sum(tmp\_liks)}\\
\STATE
\FOR{\texttt{i = 1:k}}
\STATE    \texttt{liks(i)=tmp\_liks(i)/total;}
\ENDFOR
\STATE
\RETURN \texttt{liks} \COMMENT{Array proportional to the exponentiated likelihoods}
\STATE
\end{algorithmic}
\end{algorithm}

\subsection{BMA-AODE*: Extending BMA-AODE to Sets of Probabilities}\label{sec:cma}
By BMA-AODE* we extend BMA-AODE to \emph{imprecise probabilities} \citep{walley1991}, allowing multiple specifications of the prior mass function $P(S)$; we denote the \emph{credal} set containing such prior mass functions as $\mathcal{P}(S)$. While a uniform prior represents prior \textit{indifference} between the different SPODEs, the credal set represents a condition of prior \textit{ignorance} about $S$, letting the prior probability of each SPODE vary within a large range. In principle we could let the prior probability of each SPODE vary exactly between zero and one (\emph{vacuous} model). Yet, this would generate vacuous posterior inferences, thus preventing learning from data \citep{piatti2009a}. To obtain non-vacuous posterior inferences,
we introduce non-zero lower bounds for the prior probability of the models. The resulting credal set is defined by the following constraints:
\begin{equation}\label{eq:credal-set-bma}
\mathcal{P}(S) := \left\{ P(S) \left|
\begin{array}{l}
P(s_j) \geq \epsilon \quad \forall  j=1,\ldots,k\\
\sum_{j=1}^k P(s_j) = 1
\end{array}
\right.
\right\}.
\end{equation}
The prior probability of each SPODE varies thus between $\epsilon$ and $1-(k-1)\epsilon$. The set of mass functions in Eq.(\ref{eq:credal-set-bma}) is convex;  its $k$ extreme mass functions are those assigning mass $\epsilon$ to all the SPODEs apart from a single one, to which $1-(k-1)\epsilon$ is assigned.
The constant $\epsilon$ appears in other places of this paper; in the implementation we set $\epsilon=0.01$ for all occurrences of $\epsilon$. 

The credal set in (\ref{eq:credal-set-bma}) models the fact that, before observing the data, we are ignorant about the probability of each SPODE to be the true model. Considering that $\mathcal{P}(S)$ is a set a prior mass functions, BMA-AODE* can be regarded as a set of BMA-AODE classifiers, each corresponding to a different prior. The most probable class of an instance might happen to vary, when all the different priors of the credal set are considered; in this case the classification is \textit{prior-dependent}.  When dealing with prior-dependent instances, credal classifiers \citep{corani_chapter2012,corani2008a} become \emph{indeterminate}, by returning a set of classes instead of a single class.

Before discussing how this set of classes is identified, let us introduce the concept of \textit{credal dominance} (or, for short, \textit{dominance}): class $c'$  \textit{dominates}  class $c''$ if  $c'$ is more probable than $c''$ under each prior of the credal set. If no class dominates  $c'$, then $c'$ is non-dominated.
Credal classifiers return in particular all the \textit{non-dominated} classes, identified performing different by pairwise dominance tests among classes. This criterion is called \textit{maximality} \citep[Section 3.9.2]{walley1991} and is described by Algorithm \ref{algo-non-dom}. We point the reader to \citep{troffaes2007decision} for a discussion of alternative criteria for taking decisions under imprecise probabilities.

\begin{algorithm}[!h]
\caption{Identification of the non-dominated classes $\mathcal{ND}$ through maximality}
\label{algo-non-dom}
\begin{algorithmic}
\STATE
\STATE  $\mathcal{ND}$ := $\mathcal{C}$
\STATE
\FOR {$c'\in\ \mathcal{C}$}
\FOR {$c''\in\ \mathcal{C}$ ($c'' \neq c'$)}
\STATE
\STATE  check whether $c'$ dominates $c''$
\STATE
\IF  {$c'$ dominates $c''$}
\STATE remove $c''$ from $\mathcal{ND}$
\ENDIF
\STATE
 \ENDFOR
\ENDFOR
\STATE
\RETURN $\mathcal{ND}$
\STATE
\end{algorithmic}
\end{algorithm}

Non-dominated classes are incomparable, namely there is no available information to rank them. Credal classifiers can be thus seen as dropping the dominated classes  and expressing  indecision about the non-dominated ones.

Within BMA-AODE*, $c'$ dominates $c''$ if the solution of the following optimization problem is greater than one:
\begin{eqnarray}\label{eq:maximality2}
\mbox{minimize:}\phantom{aaa}	& \displaystyle \frac{\sum_{j=1}^k P(c'|\mathbf{\tilde{a}},s_j) \cdot L_j  \cdot P(s_j) }{\sum_{j=1}^k P(c''|\mathbf{\tilde{a}},s_j)\cdot L_j   \cdot P(s_j)}  \nonumber\\
&\nonumber\\ \mbox{{subject to:}\phantom{aaa}}  & P(s_j) \geq \epsilon  \quad \forall  j=1,\ldots,k\nonumber\\
& \sum_{j=1}^k P(s_j) = 1, \nonumber
\end{eqnarray}
Note that the constrains of the problem correspond to the definition of credal set given in Eq.\ref{eq:credal-set-bma}.
The above optimization task is a \emph{fractional-linear program}; it can be mapped into a linear program by the Charnes-Cooper transformation (see Appendix \ref{app:cc}) and then solved \textit{exactly}.

As already discussed for BMA-AODE, we include in the computation only the SPODEs whose conditional likelihood is at least $L_{\mathrm{max}}/10^4$.  This can be regarded as a belief revision process, involving the credal set.  The marginal credal set $\mathcal{P}'(Y)$ induces the following revision of the joint credal set $\mathcal{P}(X,Y)$:
$$\mathcal{P}'(X,Y):= \left\{ P'(X,Y) \left| \begin{array}{l} P'(x,y):= P(x|y) \cdot P'(y) \\ P'(Y)\in\mathcal{P}'(Y) \end{array} \right. \right\}.$$
It is worth emphasizing that the prior credal of BMA-AODE* includes the uniform prior adopted by BMA-AODE;
therefore, the set of non-dominated classes identified by BMA-AODE*  includes the most probable class returned by BMA-AODE;  if in particular BMA-AODE* returns a single non-dominated class, this coincides to the class returned by BMA-AODE.

\subsection{Compression-Based averaging}\label{sec:comp}
Compression-based averaging has been introduced by \citep{boulle2007compression} as a remedy against the tendency of BMA at getting excessively concentrated around the most probable model, which indeed deteriorates the performances \citep{boulle2007compression,domingos2000bac}. This approach replaces the posterior probabilities $P(s_j|\mathcal{D})$ of the models by smoother \textit{compression} weights, which we denote as $P'(s_j|\mathcal{D})$ for model $s_j$. Note that also the adoption of the compression coefficients in place of the posterior probabilities can be seen as a belief revision.

To present the method, we need some further notation.  In particular, we denote by $LL_j$ the log of the conditional  likelihood of model $s_j$.  We moreover introduce the \textit{null classifier} as a Bayesian network with no arcs, which models the class as independent from the features and whose probabilistic classifications correspond to the marginal probabilities of the classes. The null classifier will be used for the computation of the compression coefficients.  We denote the null classifier as $s_0$; therefore we associated a further state $s_0$ to $S$, whose domain thus becomes $\{s_0,s_1,\ldots,s_k\}$.
We denote as $LL_0$ the conditional log-likelihood of the null classifier.
 It has been shown \citep{boulle2007compression} that $LL_0=-nH(C)$, where $H(C):=-\sum_{c \in \mathcal{C}} P(c)\log{P(c)}$ is the entropy\footnote{For this equivalence to hold, it is necessary computing the entropy using the natural logarithm, instead of the $\log_2$ as usual.} of the class.

Since we are dealing with a traditional single-prior classifier, we set a single prior mass function over the models, assigning uniform prior probability to the various SPODEs but prior probability $\epsilon$ to the null model; assigning a prior probability to the null model is necessary, since its posterior probability appears in the compression coefficients.  Thus, we define the prior over variable $S$ as follows:
\begin{equation}\label{eq:p}
P(s_j)=\left\{
\begin{array}{ll}
\epsilon & j=0,\\
\frac{1-\epsilon}{k}  &j=1,\ldots,k.
\end{array}
\right.
\end{equation}
The compression coefficients are computed in two steps: computation of the \textit{raw} compression coefficients and normalization. The \emph{raw} compression coefficient associated to SPODE $s_j$ is:
\begin{equation}\label{eq:unst_compr}
\pi_j:=1-\frac{\log P(s_{j}|\mathcal{D})}{\log P(s_{0}|\mathcal{D})}=1-\frac{LL_{j}+\log P(s_{j})}{LL_{0}+\log P(s_{0})}=1-\frac{LL_{j}+\log\frac{1-\epsilon}{k}}{-nH(C)+\log\epsilon},
\end{equation}
A negative $\pi_j$ means that $s_{j}$ is a worse predictor than the null model; a positive $\pi_j$ means that $s_{j}$ is a better predictor than the null model, which is the general case in practical situations. The upper limit of $\pi_j$ is one: in this case $s_j$ is a perfect predictor, with likelihood 1, and thus log-likelihood 0. Following \citep{boulle2007compression}, we keep in the ensemble only the \textit{feasible} models, namely those with $\pi_j>0$; we instead discard the models with $\pi_j<0$. Also this procedure corresponds to a belief revision induced by the removal from the ensemble of the models whose posterior probability falls below a certain threshold. Note also that, since $\pi_0=0$ by definition, the null model is not part of the resulting ensemble.  The compression coefficients can be justified  as follows \citep{boulle2007compression}: $LL_{j}+\log P(s_{j})$ ``\textit{represents the quantity of information required to encode the model plus the class values given the model. The code length of the null model can be interpreted as the quantity of information necessary to describe the classes, when no explanatory data is used to induce the model.  Each model can potentially exploit the explanatory data to better compress the class conditional information. The ratio of the code length of a model to that of the null model stands for a relative gain in compression efficiency.}''

With no loss of generality, assume the features to be ordered so that $A_1,A_2,\ldots,A_{\tilde{k}}$ yield a feasible model when used as super-parent; thus, SPODEs $s_1,s_2,\ldots,s_{\tilde{k}}$ are feasible, while SPODEs $s_j$ with $j>\tilde{k}$ are  removed from the ensemble. The \textit{normalized} compression coefficients $P'(s_j|\mathcal{D})$ are obtained by normalizing the raw compression coefficients of the feasible SPODEs:
\begin{equation}
P'(s_j|\mathcal{D}) = \left\{
\begin{array}{ll}
\frac{\pi_j}{\sum_{l=1}^{\tilde{k}} \pi_l} & \mathrm{if} \, j=1,\ldots,\tilde{k},\\
\phantom{aa} 0 & \mathrm{otherwise}.
\end{array}
\right.
\label{eq:std_compr}
\end{equation}
The posterior probabilities are estimated as:
\begin{equation}\label{eq:comp-aode}
P(c|\tilde{\mathbf{a}})=
\sum_{j=1}^k
P(c|\tilde{\mathbf{a}},s_j) \cdot P'(s_j|\mathcal{D}).
\end{equation}
We call this classifier COMP-AODE, where COMP stands for compression-based. COMP-AODE performs a weighted linear combination of probabilities estimated by different models; in risk analysis, a weighted linear combination of probabilities estimated by different experts is referred to as \textit{linear opinion pool} \citep{clemen1999combining}.

\subsection{COMP-AODE*: Extending COMP-AODE to Sets of Probabilities}\label{sec:caode}
We extend COMP-AODE to imprecise probabilities by allowing for multiple specifications of the prior $P(S)$ over the models, collected into a credal set $\mathcal{P}_c(S)$, where the subscript denotes compression. Differently from the credal set $\mathcal{P}(S)$ used by the BMA-AODE*, here we also consider the null model. We assign to the null model a fixed prior probability $\epsilon$, while the prior probability of the SPODEs are free to vary under constraints analogous to those of BMA-AODE*;  in this way we model a condition of prior \textit{ignorance}. The credal set $\mathcal{P}_c(S)$ adopted by COMP-AODE* is therefore:
\begin{equation}\label{eq:cs}
\mathcal{P}_c(S) := \left\{ P(S) \left| \begin{array}{l}
P(s_0)=\epsilon,\\
 P(s_j) \geq \epsilon \quad \forall  j=1,\ldots,k,\\
\sum_{j=0}^k P(s_j) = 1
\end{array}
\right.
\right\}.
\end{equation}

The bounds of the raw compression coefficient defined in Eq.(\ref{eq:unst_compr})  are obtained by letting vary $P(S)$ in $\mathcal{P}_c(S)$:
\begin{equation}\label{eq:impr-compr}
\pi_{j}\in \left[1-\frac{LL_{j}+\log\epsilon}{-nH(C)+\log\epsilon},\,\,1-\frac{LL_{j}+\log\left(1-k\epsilon\right)}{-nH(C)+\log\epsilon}\right].
\end{equation}
Since the prior used by COMP-AODE (Eq. \ref{eq:p}) belongs to the credal set of COMP-AODE*,  the point estimate of the compression coefficient adopted by COMP-AODE (Eq.\ref{eq:unst_compr}) lies in the above interval. Note also that the upper bound of the above interval (\textit{upper coefficient of compression}) is obtained in correspondence of the \textit{extreme} mass function which assigns prior probability $1-k\epsilon$ to model $s_j$ and prior probability $\epsilon$ to all the remaining models. The various $\pi_{j}$ cannot vary independently from each other; they are instead linked by the normalization constraint in Eq.(\ref{eq:cs}).

COMP-AODE* regards SPODE $s_j$ as non-feasible if the \emph{upper} coefficient of compression is non-positive: this approach thus preserves all the models which are feasible, in the sense of Section \ref{sec:comp},  for at least a prior in the set $\mathcal{P}_c(S)$. COMP-AODE* is thus more conservative than COMP-AODE, namely it discards a lower number of models. However, generally neither COMP-AODE* nor COMP-AODE remove any SPODE from the ensemble.
Since the prior adopted by COMP-AODE is contained in the credal set of COMP-AODE*, the most probable class identified by COMP-AODE is part of the non-dominated classes identified by COMP-AODE*. \footnote{Exception to this statement are in principle possible if the set of feasible SPODEs differs between
COMP-AODE*  and COMP-AODE. However, this did not happen in our extensive experiments. }

Like BMA-AODE*, COMP-AODE* identifies for each instance the non-dominated classes through maximality (Algorithm \ref{algo-non-dom}).
In the following, we explain how to compute the test of dominance among two classes.

\subsubsection*{Testing dominance}
Without loss of generality, we assume the features to have be re-ordered, so that the first $\tilde{k}$ features yield a model with positive \textit{upper} coefficient of compression when used as super-parent; in other words,  SPODEs $\{s_1,\ldots,s_{\tilde{k}}\}$ are the feasible ones. In this case the dominance test corresponds to evaluate whether or not the solution of the following optimization problem is greater than one.

\begin{eqnarray}\label{eq:maximality2}
	\mbox{minimize:}\phantom{aaa}	& \displaystyle \frac{P(c'|\mathbf{a})}{P(c''|\mathbf{a})} \propto \frac{\sum_{j=1}^{\tilde{k}} P(c'|\mathbf{\tilde{a}},s_j) \cdot \pi_j}{\sum_{j=1}^{\tilde{k}} P(c''|\mathbf{\tilde{a}},s_j)\cdot \pi_j }  \nonumber\\
\\
\mbox{w.r.t.:} \phantom{aaa} & \displaystyle P(s_0),P(s_1),\ldots,P(s_k)\\
	&\nonumber\\ \mbox{{subject to:}\phantom{aaa}}& P(s_0)=\epsilon\\  & P(s_j) \geq \epsilon  \quad \forall  j=1,\ldots,k\nonumber\\
& \sum_{j=1}^k P(s_j) = 1, \nonumber
\end{eqnarray}
where the normalization term $\sum_{j=1}^{\tilde{k}} \pi_j$ has been already simplified, being positive by definition.
Recalling that $P(s_0)=\epsilon$ and introducing Eq.(\ref{eq:unst_compr}) which shows how 
$\pi_j$ depends on the optimization variable $P(s_j)$, we rewrite the objective function as:
$$\frac{\sum_{j=1}^{\tilde{k}}P(c'|\mathbf{a},s_j)\cdot \left(1-\frac{\log P(s_{j})+LL_{j}}{\log\epsilon+LL_{0}}\right)}{\sum_{j=1}^{\tilde{k}}P(c''|\mathbf{a},s_j)\cdot \left(1-\frac{\log P(s_{j})+LL_{j}}{\log\epsilon+LL_{0}}\right)},$$
and hence
$$\frac{\sum_{j=1}^{\tilde{k}}P(c'|\mathbf{a},s_j)\cdot \left(\log\epsilon+LL_{0}-LL_{j}\right)-\sum_{j=1}^{\tilde{k}}P(c'|\mathbf{a},s_j)\cdot \log P(s_j)}{\sum_{j=1}^{\tilde{k}}P(c''|\mathbf{a},s_j)\cdot \left(\log\epsilon+LL_{0}-LL_{j}\right)-\sum_{j=1}^{\tilde{k}}P(c''|\mathbf{a},s_j)\cdot \log P(s_j)}.$$
We then introduce the constants $a:=\sum_{j=1}^{\tilde{k}}P(c'|\mathbf{a},s_j)\left(\log\epsilon+LL_{0}-LL_{j}\right)$,
$b:=\sum_{j=1}^{j=\tilde{k}}P(c''|\mathbf{a},s_j)\left(\log\epsilon+LL_{0}-LL_{j}\right)$,
$\alpha_{j}:=P(c'|\mathbf{a},s_j)$, $\beta_{j}:=P(c''|\mathbf{a},s_j)$. 
After changing the sign of both numerator and denominator of the objective function, we rewrite the optimization problem, with respect to the variables ${x_{1},x_{2},\ldots,x_{\tilde{k}}}$, where $x_{j}:=\log P(s_{j})$, as follows:
\begin{eqnarray*}
\mbox{minimize:}\phantom{aaa}	& \displaystyle \frac{\sum_{j=1}^{\tilde{k}}\alpha_{j} x_{j} -a}{\sum_{j=1}^{\tilde{k}}\beta_{j}x_{j}-b} \nonumber \\
\\
\mbox{w.r.t.:} \phantom{aaa} & \displaystyle x_1,\ldots,x_{\tilde{k}}\\
&\nonumber\\ \mbox{{subject to:}\phantom{aaa}}&  x_j \geq  \log\epsilon \,\ \forall j=1,\ldots,\tilde{k},\\
&\sum_{j=1}^{\tilde{k}}\exp{x_{j}}=1-\epsilon-(k-\tilde{k})\epsilon. \nonumber
\end{eqnarray*}
where the constrains are derived from the definition of credal set (\ref{eq:cs}). The last constraint is justified as follows: $(k-\tilde{k})$ models have been removed from the ensemble as unfeasible and therefore they do not appear in the optimization problem.  Without changing the credal set, we set their priors to $\epsilon$; since these models do not impact on the objective function, the best solution is attained by allocating to them the minimum possible prior probability.  We then substitute $y_j:=\exp{x_j}$ to avoid numerical problems in the optimization, thus getting the following non-linear optimization problem with linear constraints:
\begin{eqnarray}\label{eq:maximality2}
\mbox{minimize:}\phantom{aaa}	& \displaystyle \frac{\sum_{j=1}^{\tilde{k}}\alpha_{j} \cdot \log y_{j} -a}{ \sum_{j=1}^{\tilde{k}}\beta_{j}\cdot \log y_j -b} \nonumber \\ \\
\mbox{w.r.t.:} \phantom{aaa} & \displaystyle y_1,\ldots,y_{\tilde{k}}\\
&\nonumber\\ \mbox{{subject to:}\phantom{aaa}}& y_j \geq \epsilon,\\  
& \sum_{j=1}^{\tilde{k}}  y_j = 1 - (k -\tilde{k} - 1) \epsilon. \nonumber
\end{eqnarray}

\subsection{Computational Complexity of the Classifiers}
We now analyze the computational complexity of the proposed classifiers and compare it with that of the standard AODE.  We distinguish between \emph{learning} and \emph{classification} complexity, the latter referring to the classification of a single instance. Both the \emph{space} and the \emph{time} required for computations are evaluated. The orders of magnitude of these descriptors are reported as a function of the dataset size $n$, the number of attributes/SPODEs $k$, the number of classes $l:=|\mathcal{C}|$, and average number of states for the attributes $v:=k^{-1}\sum_{i=1}^k|\mathcal{A}_i|$. A summary of this analysis is in Table \ref{tab:complex} and the discussion below.

\begin{table}[!h]
	\begin{centering}
		\begin{tabular}{lcccc}\toprule
% 			\hline
% 			\hline
			Algorithm& Space&\multicolumn{2}{c}{Time}\\ 
			&learning/classification & learning & classification\\ \midrule
% 			\hline
			AODE        & $\mathcal{O}(lk^2v^2)$  &  $\mathcal{O}(nk^2)$     & $\mathcal{O}(l k^2)$\\
			BMA-AODE/COMP-AODE  & $\mathcal{O}(lk^2v^2)$  &  $\mathcal{O}(n(l+k)k)$  & $\mathcal{O}(l k^2)$\\
			BMA-AODE*/COMP-AODE* & $\mathcal{O}(lk^2v^2)$  &  $\mathcal{O}(n(l+k)k)$  & $\mathcal{O}(l^2 k^3)$\\ \bottomrule
% 			\hline
		\end{tabular}
		\par
	\end{centering}
	\caption{Complexity of classifiers.}
	\label{tab:complex}
\end{table}

Let us first evaluate the AODE. For a single SPODE  $s_j$, the tables $P(C)$, $P(A_j|C)$ and $P(A_i|C,A_j)$, with $i=1,\ldots,k$ and $i \neq j$ should be stored, this implying space complexity $\mathcal{O}(lkv^2)$ for learning each SPODE and $\mathcal{O}(lk^2v^2)$ for the AODE ensemble. These tables should be available during learning and classification for both classifiers; thus, space requirements of these two stages are the same.  

Time complexity to scan the dataset and learn the probabilities is $\mathcal{O}(nk)$ for each SPODE, and hence $\mathcal{O}(nk^2)$ for the AODE. The time required to compute the posterior probabilities is $\mathcal{O}(lk)$ for each SPODE, and hence $\mathcal{O}(lk^2)$ for AODE.

Learning BMA-AODE or COMP-AODE takes the same space as AODE, but higher computational time, due to the evaluation of the conditional likelihood as in Eq.(\ref{eq:cond_lik}). The additional computational time is  $\mathcal{O}(nlk)$, thus requiring $\mathcal{O}(n(l+k)k)$ time overall. For classification, time and space complexity during learning and classification are just the same.

The credal classifiers BMA-AODE* and COMP-AODE* require  the same space complexity and the same time complexity in learning
of their non-credal counterparts. However, credal classifiers have higher time complexity in classification. The pairwise dominance tests in Algorithm \ref{algo-non-dom} requires the solution of a number of optimization problems for each test instance which is quadratic in the number of classes.
We can roughly describe as cubic in the number of variables the time complexity of solving the linear programming problem for BMA-AODE* and the optimization of the non-linear function, with linear constraints, for COMP-AODE*.
Summing up credal classifiers increase of one unit, compared to their single-prior counterparts, the exponents of the number of classes and attributes
in the time complexity of the classification stage.

\section{Experiments}\label{sec:exp}
We run experiments on 40 data sets, whose characteristics are given in the Appendix (Table \ref{tab:dsets}). On each data set we perform 10 runs of 5-fold cross-validation. In order to have complete data, we replace missing values with the median and the mode for respectively numerical and categorical features.  We discretize numerical features by the entropy-based method of \citep{fayyad1993}.  For pairwise comparison of of classifiers over the collection of data sets we use the non-parametric Wilcoxon signed-rank test.\footnote{For each indicator of performance, we generate two \textit{paired} vectors: the same position in both vectors refers to the same data set. The two vectors are then used as input for the test.} The Wilcoxon signed-rank test is indeed recommended for comparing two classifiers on multiple data sets \citep{demsar06jmlr}: being non-parametric it avoids strong assumptions and robustly deals with outliers.

\subsection{Determinate classifiers}
We call \textit{determinate} the classifiers which always return a single class, namely AODE, BMA-AODE and COMP-AODE.  For determinate classifiers we use two indicators: the accuracy, namely the percentage of correct classifications, and the Brier loss
$$
\frac{1}{n_{te}}\sum_{i}^{n_{te}} \left(1-P(c^{(i)}|\mathbf{a}^{(i)})\right)^2,
$$
where $n_{te}$ denotes the number of instances in the test set, while $P(c^{(i)}|\mathbf{a}^{(i)})$ is the probability estimated by the classifier for the true class of the $i$-th instance. The Brier loss assesses the quality of the estimated probabilities in a more sensitive way than accuracy.

A first finding is that AODE outperforms BMA-AODE, having both higher accuracy ($p$-value $<$ .01) and lower Brier loss.  We present in Figure \ref{subfig:bma-acc} the scatter plot of accuracies and in Figure \ref{subfig:bma-brier} the \textit{relative} Brier losses, namely  the Brier loss of BMA-AODE divided, data set by data set, by the Brier loss of AODE. On average, BMA-AODE  has 3\%  higher Brier loss than AODE.  The fact that AODE  outperforms BMA-AODE could be expected; the same finding was already given in \citep{yang2007select} and in \citep{cerquides2005robust}, with the main difference that  the BMA-AODE of these works was based on the marginal likelihood rather than on the conditional likelihood.  Our results show that BMA-AODE is outperformed by AODE, even when using the conditional likelihood. BMA-AODE is outperformed by AODE both because  its excessive concentration around the most probable model \citep{boulle2007compression,cerquides2005robust,domingos2000bac,minkaBMA} which tends to cancel the advantage of averaging over models, and  because of the effectiveness of simply averaging over SPODEs, as done by AODE, in terms of reduction of the variance error.

As outlined by Figure \ref{subfig:comp-acc}, the difference between accuracies is instead not significant when comparing COMP-AODE and AODE.  However COMP-AODE outperforms AODE on the Brier loss ($p$-value $<$ .01); in Figure \ref{subfig:comp-brier} we show the \textit{relative} Brier losses, namely the Brier loss of COMP-AODE divided, data set by data set, by the Brier loss of AODE.  Averaging over data sets, COMP-AODE reduces the Brier loss of about 3\% compared to AODE. We see this result as noteworthy, since AODE is a high performance classifier. These positive results  with the compression-based approach broaden the scope of the experiments of \citep{boulle2007compression}, in which the compression approach was applied to an ensemble of naive Bayes classifiers.
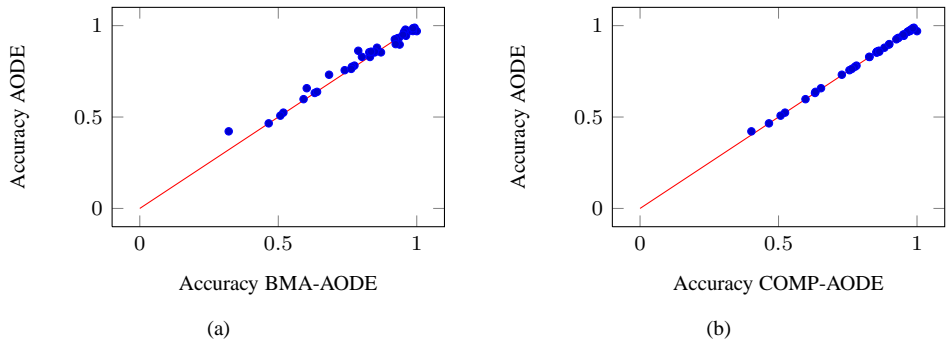
\begin{figure}[!h]
\centering
 \centerline{\hbox{
\subfigure[]{
\label{subfig:bma-acc}
\begin{tikzpicture}[]
\begin{axis}[width=6cm,height=4.5cm, ylabel=Accuracy AODE, xlabel=Accuracy BMA-AODE]
% \addplot +[only marks]  table[col sep=comma, x=acc_DBMA-AODE, y=acc_AODE] {results_aode.csv};
\addplot +[only marks]  table[col sep=comma, x=acc_DBMA-AODE, y=acc_AODE] {results_aode.tex};
\addplot  +[mark=none] plot coordinates { (0,0) (1,1)};
\end{axis}
\end{tikzpicture}
}
\hspace{0.4cm}
\subfigure[]{
\label{subfig:comp-acc}
\begin{tikzpicture}
\begin{axis}[width=6cm,height=4.5cm,  ylabel=Accuracy AODE, xlabel=Accuracy COMP-AODE]
% \addplot+[only marks]  table[col sep=comma,  x=acc_C-DBMA-AODE, y=acc_AODE] {results_aode.csv};
\addplot+[only marks]  table[col sep=comma,  x=acc_C-DBMA-AODE, y=acc_AODE] {results_aode.tex};
\addplot +[mark=none] plot coordinates { (0,0) (1,1)};
\end{axis}
\end{tikzpicture}
}}}
\caption{\label{fig:acc-scatter}Scatter plots of accuracies; the solid line shows the bisector.}
\end{figure}
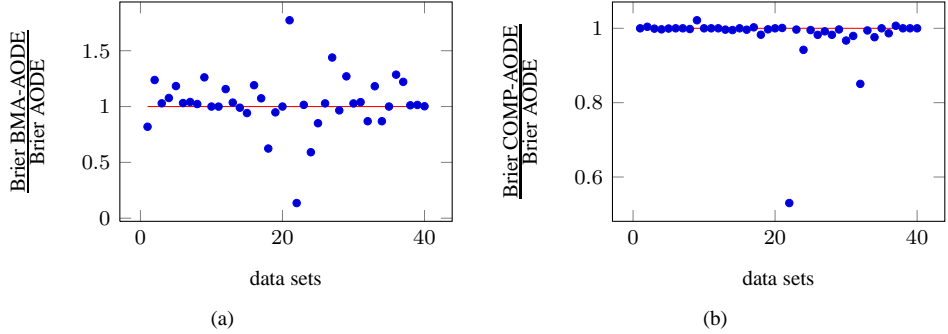
\begin{figure}[!h]
\centering
\vspace{.5cm}
\centerline{\hbox{
\subfigure[]{
\label{subfig:bma-brier}
\begin{tikzpicture}
\begin{axis}[width=6cm,height=4.5cm, xlabel= data sets, ylabel=$\frac{\mbox{Brier BMA-AODE}}{\mbox{Brier AODE}}$]
% \addplot+[only marks]   table[col sep=comma, y expr=\thisrow{br_score_DBMA}/\thisrow{br_score_AODE}] {results_aode.csv};
\addplot+[only marks]   table[col sep=comma, y expr=\thisrow{br_score_DBMA}/\thisrow{br_score_AODE}] {results_aode.tex};
\addplot  +[mark=none]plot coordinates { (1,1) (40,1)};;
\end{axis}
\end{tikzpicture}
}
\hspace{0.2cm}
\subfigure[]{
\label{subfig:comp-brier}
\begin{tikzpicture}
\begin{axis}[width=6cm,height=4.5cm, xlabel= data sets, ylabel=$\frac{\mbox{Brier COMP-AODE}}{\mbox{Brier AODE}}$]
% \addplot+[only marks]   table[col sep=comma, y expr=\thisrow{br_score_C-DBMA-AODE}/\thisrow{br_score_AODE}] {results_aode.csv};
\addplot+[only marks]   table[col sep=comma, y expr=\thisrow{br_score_C-DBMA-AODE}/\thisrow{br_score_AODE}] {results_aode.tex};
\addplot  +[mark=none] plot coordinates { (1,1) (40,1)};
\end{axis}
\end{tikzpicture}
}}}
\caption{\label{fig:acc-brier}Relative Brier losses; points lying \textit{below} the horizontal line represent performance better than AODE, and vice versa.
Note the different y-scales of the two graphs. }
\end{figure}

\subsection{Credal classifiers}
A credal classifier can be seen as separating the instances into two groups: the \textit{safe} ones, for which it returns a single class is returned, and the \textit{prior-dependent} ones, for which it returns two or more classes. Note that prior-dependence is not an intrinsic property to the instance: an instance can be judged as prior-dependent by a certain credal classifier and as safe by a different credal classifier.
To characterize the performance of a credal classifier, the following four indicators are considered \citep{corani2008a}:
%Thus, prior-dependence is a characteristic of a certain instance, when analyzed by a \textit{specific} credal classifier.
\begin{itemize}
\item \textit{determinacy}: \% of instances recognized as safe, namely classified with a single class;
\item \textit{single-accuracy}: the accuracy achieved over the instances recognized as safe;
\item \textit{set-accuracy}: the accuracy achieved, by returning a set of classes, over the prior-dependent instances;
\item \textit{indeterminate output size}: the average number of classes returned on the prior-dependent instances.
\end{itemize}

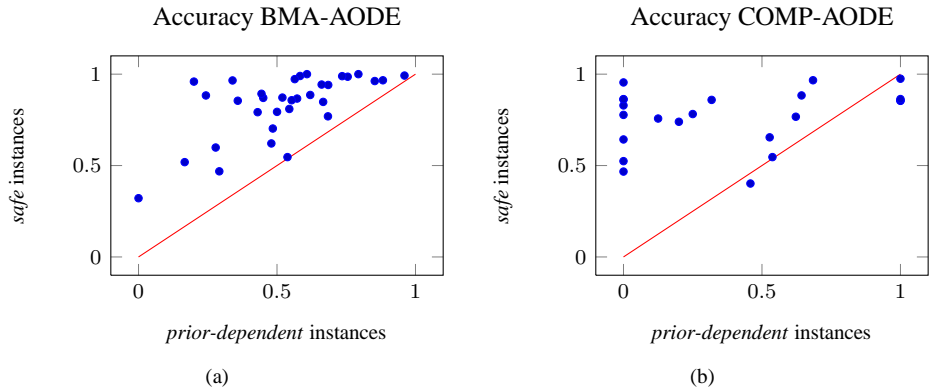
\begin{figure}[!ht]
\vspace{.5cm}
\centerline{\hbox{
\subfigure[]{
\label{subfig:drop-bma}
\begin{tikzpicture}
\begin{axis}[width=6cm,height=4.5cm, xlabel= \textit{prior-dependent} instances, title=Accuracy BMA-AODE, ylabel=\textit{safe} instances]
% \addplot+[only marks]  table[col sep=comma, x=precise_set_DBMA-AODE*, y=precise_single_DBMA-AODE*] {results_aode.csv};
\addplot+[only marks]  table[col sep=comma, x=precise_set_DBMA-AODE*, y=precise_single_DBMA-AODE*] {results_aode.tex};
\addplot +[mark=none]plot coordinates { (0,0) (1,1)};
\end{axis}
\end{tikzpicture}
}
\hspace{0.2cm}
\subfigure[]{
\label{subfig:drop-comp}
\begin{tikzpicture}
\begin{axis}[width=6cm,height=4.5cm,xlabel= \textit{prior-dependent} instances, title=Accuracy COMP-AODE, ylabel=\textit{safe} instances]
% \addplot+[only marks]  table[col sep=comma, x=precise_set_C-DBMA-AODE*, y=precise_single_C-DBMA-AODE*] {results_aode.csv};
\addplot+[only marks]  table[col sep=comma, x=precise_set_C-DBMA-AODE*, y=precise_single_C-DBMA-AODE*] {results_aode.tex};
\addplot +[mark=none] plot coordinates { (0,0) (1,1)};
\end{axis}
\end{tikzpicture}
}}}
\caption{\label{fig:scatter-drops}Accuracy of the determinate classifiers on the instances recognized as safe and as prior-dependent by their credal counterparts. The accuracies of BMA-AODE [COMP-AODE] is thus separately measured on the  instances judged safe and prior-dependent by BMA-AODE* [COMP-AODE*]. The solid line shows the bisector.}
\end{figure}

Averaging over data sets, BMA-AODE*  has 94\% determinacy; it is completely determinate on 7 data sets.  However, this determinacy fluctuates among data sets, showing for instance a significant correlation with the sample size $n$ ($\rho=0.3$).  The choice of the prior is less important on large data sets: bigger data sets tend to contain a lower percentage of prior-dependent instances, thus increasing determinacy. BMA-AODE* performs well when indeterminate: averaging over all data sets, it achieves 90\% set-accuracy by returning 2.3 classes (the average number of classes in the collection of data sets is 3.6).  It is worth analyze the performance of BMA-AODE on the prior-dependent instances.  In Figure \ref{subfig:drop-bma} we compare, data set by data set, the accuracy achieved by BMA-AODE on the instances judged respectively as safe and as prior-dependent by BMA-AODE*; the plot shows a sharp drop of accuracy on the prior-dependent instances, which is  statistically significant ($p$-value $<$ .01).  As a rough indication, averaging over data sets, the accuracy of BMA-AODE is 83\% on the safe instances but only 52\% on the instances recognizes as prior-dependent by BMA-AODE*.  Thus, on the prior-dependent instances, BMA-AODE provides fragile classifications; on the same instances, BMA-AODE* returns a small-sized but highly accurate set of classes.

Let us now analyze the performance of COMP-AODE*; it has higher determinacy than BMA-AODE*;  averaging over data sets, its determinacy is 99\%, with only minor fluctuations across data sets; the classifier is moreover completely determinate on 18 data sets.  The determinacy of COMP-AODE* is very high and stable across data sets. Therefore, under the compression-based approach only a small fraction of the instances is prior-dependent; this robustness to the choice of the prior is likely to contribute to the good performance of compression-based ensemble of classifiers and constitutes a desirable but previously unknown property of the compression-based approach. Numerical inspection shows that the logarithmic smoothing of the models' posterior probabilities makes indeed the compression weights only little sensitive to the choice of the prior. COMP-AODE* performs well when indeterminate: averaging over all data sets, it achieves 95\% set-accuracy by returning 2 classes (note that the indeterminate output size cannot be less than two).  

Again, it is worth checking the behavior of the corresponding determinate classifier, namely COMP-AODE, on the instances that are prior-dependent for the  COMP-AODE*. In Figure \ref{subfig:drop-comp} we compare, data set by data set, the accuracy achieved by COMP-AODE on the instances judged respectively safe and prior-dependent by COMP-AODE*; there is a large drop of accuracy on the prior-dependent instances, and  the drop is significant ($p$-value $<$ .01).  Averaging over data sets, the accuracy of COMP-AODE drops from  82\% on the safe instances to only 47\% on the instances judged as prior-dependent by COMP-AODE*. Even COMP-AODE, despite its robustness to the specification of the prior, undergoes a severe loss of accuracy on the instances recognized as prior-dependent by COMP-AODE*. On the very same instances, COMP-AODE* returns a small sized but highly reliable set of classes, thus enhancing the overall classification reliability.

\subsection{Utility-based Measures}
We have seen so far that the credal classifiers extend in a sensible way their determinate counterparts, being able to recognize prior-dependent instances and to robustly deal with them. Yet, it is not obvious how to compare credal and determinate classifiers by means of a synthetic indicator.  In fact, to fairly compare determinate and indeterminate predictions is very challenging; to the best of our knowledge, a satisfactory solution exists only for 0-1 loss, while comparing
determinate and indeterminate predictions in a cost-sensitive setting, in which different kind of errors imply different costs, is still an open problem. In the following we thus reason under 0-1 loss.
The \emph{discounted accuracy}  rewards a prediction made of $m$ classes with $1/m$ if it contains the true class, and with 0 otherwise. Discounted accuracy is then compared to the accuracy achieved by a determinate classifier. A theoretical justification for discounted-accuracy  has been given by \cite{2011:43:isipta} showing that,  within a betting framework based on fairly general assumptions, discounted-accuracy is the only score which satisfies some fundamental properties for assessing both determinate and indeterminate classifications. Yet \cite{2011:43:isipta} also shows some severe limits of discounted-accuracy, which we illustrate by means of an example: we consider two different medical doctors, doctors \textit{random} and doctor \textit{vacuous}, who should decide whether a patient is \emph{healthy} or \emph{diseased}.  Doctor \textit{random} issues random diagnosis,  using a uniform distribution over the two categories.  Doctor \textit{vacuous} instead always return both categories, admitting to be ignorant.  Let us assume that the hospital profits a quantity of money  proportional to the discounted-accuracy achieved by its doctors at each visit.  Both doctors have the same \textit{expected} discounted-accuracy for each visit,  namely $1/2$.  For the hospital, both doctors provide the same \textit{expected} profit on each visit, but with a substantial difference: the profit of doctor vacuous is \textit{deterministic}, while the profit of doctor random is affected by considerable variance.  Any risk-averse hospital manager should thus prefer doctor vacuous over doctor random, since it yields the same expected profit  with less variance. In fact, under risk-aversion, the expected utility increases with expectation of the rewards  and decreases with their variance \citep{levy1979approximating}.
% The higher reliability of doctor vacuous in earning rewards is thus reflected by a smaller variance of discounted accuracy.
To capture this point it is necessary  introducing a utility function, to be then applied on the discounted-accuracy score assigned on each instance.
In \cite{2011:43:isipta} the utility function is designed as follows: the utility of a correct and determinate classification (discounted-accuracy 1) is 1; the utility of a wrong classification (discounted-accuracy  0) is 0; the utility of an accurate but indeterminate classification consisting of two classes (discounted-accuracy 0.5)  is assumed to lie between 0.65 and 0.8. Two quadratic utility functions are then derived corresponding to these boundary values, and 
passing respectively  through $\{u(0)=0,u(0.5)=0.65,u(1)=1\}$ and  $\{u(0)=0,u(0.5)=0.8,u(1)=1\}$, denoted as $u_{65}$ and $u_{80}$ respectively\footnote{The mathematical expression of these utility functions are as follows: $u_{65}(x)=-1.2x^2 + 2.2x$, $u_{80}(x)=-0.6x^2 + 1.6x$, where $x$ is the value of discounted accuracy.}.
% The utility functions are applied on the discounted-accuracy score obtained by the classifier on each instance.
Since $u(1)=1$, utility and accuracy coincide for determinate classifiers; therefore, utility of credal classifiers and accuracy of determinate classifiers
can be directly compared.
In \cite{JdalCozlearning}  classifiers which return indeterminate classifications are scored through the $F_1$-metric, originally designed for Information Retrieval tasks. The $F_1$ metric, when applied to indeterminate classifications, returns a score which is always comprised between $u_{65}$ and $u_{80}$, further confirming the reasonableness of both utility functions. More details on the links between $F_1$, $u_{65}$ and $u_{80}$  are given in \cite{zaffalonTech2012}. We remark that in real applications the utility function should be elicited by discussion with the decision maker;
in this paper we use $u_{65}$ and $u_{80}$ to model two reasonable but different degrees of risk-aversion.

\begin{figure}[!ht]
\centering
\vspace{.5cm}
\centerline{\hbox{
\subfigure{
\begin{tikzpicture}
\begin{axis}[width=6cm,height=4.5cm, xlabel= datasets, ymax=1.25, ylabel=$\frac{u_{65}\mbox{BMA-AODE*}}{\mbox{Accuracy BMA-AODE}}$]
% \addplot+[only marks]  table[col sep=comma, x=idx, y=rel_u65bma] {results_aode.csv};
\addplot+[only marks]  table[col sep=comma, x=idx, y=rel_u65bma] {results_aode.tex};
\addplot +[mark=none] plot coordinates { (1,1) (40,1)};;
\end{axis}
\end{tikzpicture}
}
\hspace{0.4cm}
\subfigure{
\begin{tikzpicture}
\begin{axis}[width=6cm,height=4.5cm, xlabel= datasets, ymax=1.25, ylabel=$\frac{u_{80}\mbox{BMA-AODE*}}{\mbox{Accuracy BMA-AODE}}$]
% \addplot+[only marks]  table[col sep=comma, x=idx, y=rel_u80bma] {results_aode.csv};
\addplot+[only marks]  table[col sep=comma, x=idx, y=rel_u80bma] {results_aode.tex};
\addplot +[mark=none] plot coordinates { (1,1) (40,1)};;
\end{axis}
\end{tikzpicture}}}}
\vspace{.5cm}
\centerline{\hbox{
\subfigure{
\begin{tikzpicture}
\begin{axis}[width=6cm,height=4.5cm, xlabel= datasets, ymax=1.045, ylabel=$\frac{u_{65}\mbox{COMP-AODE*}}{\mbox{accuracy COMP-AODE}}$]
% \addplot+[only marks]   table[col sep=comma, x=idx, y=rel_u65comp] {results_aode.csv};
\addplot+[only marks]   table[col sep=comma, x=idx, y=rel_u65comp] {results_aode.tex};
\addplot  +[mark=none]plot coordinates { (1,1) (40,1)};;
\end{axis}
\end{tikzpicture}
}
\hspace{0.4cm}
\subfigure{
\begin{tikzpicture}
\begin{axis}[width=6cm,height=4.5cm, xlabel= datasets, ymax=1.045,ylabel=$\frac{u_{80s}\mbox{COMP-AODE*}}{\mbox{accuracy COMP-AODE}}$]
% \addplot+[only marks]  table[col sep=comma, x=idx, y=rel_u80comp] {results_aode.csv};
\addplot+[only marks]  table[col sep=comma, x=idx, y=rel_u80comp] {results_aode.tex};
\addplot  +[mark=none] plot coordinates { (1,1) (40,1)};;
\end{axis}
\end{tikzpicture}
}}}
\caption{\label{fig:relative-util}Relative utilities of credal classifiers compared to their precise counterparts.}
\end{figure}
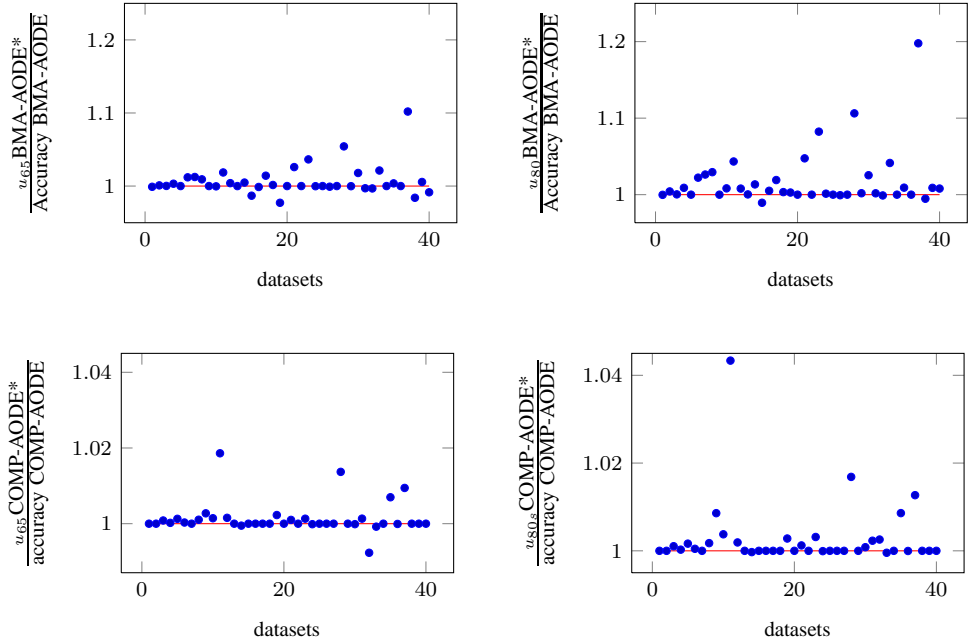

We now analyze the utilities  generated by the various classifiers, comparing each credal classifier with its determinate counterpart.  BMA-AODE*  has significantly higher utility ($p$-value $<$ .01)  than BMA-AODE under both $u_{65}$ and $u_{80}$.  This confirms that extending the model to imprecise probability is a sensible approach.  In the first row of  Figure \ref{fig:relative-util} we show the \textit{relative} utility, namely the utility of BMA-AODE* divided, data set by data set, by the utility (i.e., accuracy) of BMA-AODE; the two plots refer respectively to $u_{65}$ and $u_{80}$. Averaging over data sets, the improvement of utility is about 1\% and 2\% under $u_{65}$ and $u_{80}$; although the improvement might look small, we recall that it is obtained by  modifying the classifications of the prior-dependent instances only, 6\% of the total on average. If we focus on the prior-dependent instances only, the increase of utility generally varies between +10\% and +40\% depending on the data set and on the utility function. Clearly, the improvement is even larger under $u_{80}$ which assigns higher utility than $u_{65}$ to the indeterminate but accurate classifications.
  
The analysis is similar when comparing COMP-AODE* with COMP-AODE. In the second row of Figure \ref{fig:relative-util} we show the \textit{relative} utility, namely the utility of COMP-AODE* divided, data set by data set, by the utility (i.e., accuracy) of COMP-AODE.  The increase of utility is in this case generally under 1\%, as a consequence of the higher determinacy of COMP-AODE (99\% on average), which allows less room for improving utility through indeterminate classifications. In fact, the robustness of COMP-AODE  to the choice of the prior reduces the portion of  instances where it is necessary making the classification indeterminate.  Focusing however  on the (rare) indeterminate instances, the increase of utility deriving to the extension to imprecise probability lies between 39\% and 60\%, depending on the data set and on the utility function. Eventually, COMP-AODE* has significantly  ($p$-value $<$ .01) higher utility  than COMP-AODE under \textit{both} $u_{65}$ and $u_{80}$; also in this case the extension to the credal paradigm is beneficial.

The utilities of COMP-AODE* and BMA-AODE* are also compared; under $u_{65}$ COMP-AODE* yields significantly ($p$-value $<$ .05) higher utility than  BMA-AODE*, while under $u_{80}$ the difference among the two classifiers is not significant, although  the utility generated by COMP-AODE* is generally slightly higher. The point is that BMA-AODE*  is more often indeterminate than COMP-AODE*; under $u_{80}$ the indeterminate but accurate classifications are rewarded more than under  $u_{65}$,  thus allowing BMA-AODE* to almost close the gap with COMP-AODE*. We conclude however that COMP-AODE* should be generally preferred over BMA-AODE*.

Eventually we point out that COMP-AODE*  generates significantly ($p$-value $<$ .01)  higher utility than AODE, under \textit{both} $u_{65}$ and $u_{80}$. The extension to imprecise probability has thus concretely improved the overall performance of the compression-based ensemble: recall that
the determinate COMP-AODE yields better probability estimates but not better accuracy than AODE.
% \newpage
\subsection{Comparison with previous credal classifiers}
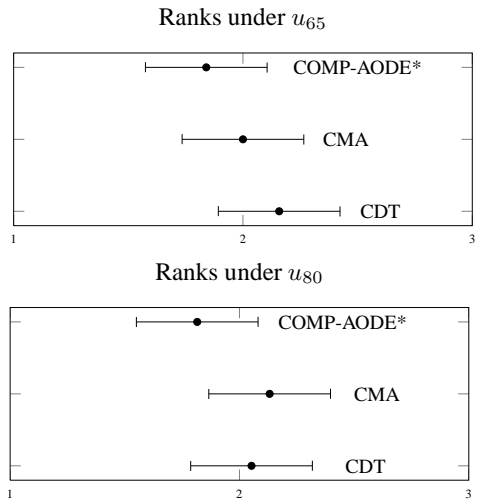
\begin{figure}[!ht]
\centering
\begin{tikzpicture}[scale=.95]
\begin{axis}[xmin=1,xmax=3,xlabel={},width=8cm,height=4cm,ytick={1,2,3},yticklabels={,,},xtick={1,2,3,4},xticklabels={\tiny 1,\tiny 2,\tiny 3,\tiny 4}, title=Ranks under $u_{65}$]
% \addplot[mark=*,only marks,error bars/x dir=both, error bars/x fixed=0.2653]coordinates { (3.346154,1)(2.987179,2)(2.666667,3)};
\addplot[mark=*,only marks,error bars/x dir=both, error bars/x fixed=0.2653]coordinates { (2.1579,1)(2,2)(1.84,3)};
\node at (axis cs:2.6,1) {\footnotesize CDT};
\node at (axis cs:2.45,2) {\footnotesize CMA};
\node at (axis cs:2.5,3) {\footnotesize COMP-AODE*};
\end{axis}
\end{tikzpicture}
\begin{tikzpicture}[scale=.95]
%\begin{axis}[xmin=1,xmax=4,xlabel={},width=5cm,height=2.5cm,ytick={1,2,3},yticklabels={,,},xtick={1,2,3,4},xticklabels={\tiny 1,\tiny 2,\tiny 3,\tiny 4}]
\begin{axis}[xmin=1,xmax=3,xlabel={},width=8cm,height=4cm,ytick={1,2,3},yticklabels={,,},xtick={1,2,3,4},xticklabels={\tiny 1,\tiny 2,\tiny 3,\tiny 4}, title=Ranks under $u_{80}$]
\addplot[mark=*,only marks,error bars/x dir=both, error bars/x fixed=0.2653]coordinates {(2.0526,1)(2.1316,2)(1.8158,3)};
\node at (axis cs:2.55,1) {\footnotesize CDT};
\node at (axis cs:2.6,2) {\footnotesize CMA};
\node at (axis cs:2.45,3) {\footnotesize COMP-AODE*};
\end{axis}
\end{tikzpicture}
\label{fig:compcred}
\caption{Comparison between credal classifiers by means of the Friedman test: the boldfaced points show the average ranks; a lower rank implies better performance. The bars display the critical distance, computed with 95\% confidence: the performance of two classifiers are significantly different  if their bars
do not overlap.}
\end{figure}

In this section we compare COMP-AODE*  with previous credal classifiers.
A well-known credal  classifier is the \emph{naive credal classifier} (NCC) \citep{corani2008a}, which is an extension of naive Bayes to imprecise probability.
We have ran NCC on the same collection of data sets following the experimental setup of Section~\ref{sec:exp}; under \textit{both} $u_{65}$ and $u_{80}$, the utility produced by
COMP-AODE* is significantly higher ($p<$0.01) than that produced by NCC. 
Thus, COMP-AODE* outperforms NCC.

However, over time algorithms more sophisticated than NCC have been developed, such as:
\begin{itemize}
 \item \emph{credal model averaging} (CMA) \citep{corani2008d}, namely a generalization of BMA (in the same spirit of BMA-AODE) for naive Bayes classifier;
\item \textit{credal decision tree} (CDT) \citep{abellan2005a}, namely  an extension of classification trees to imprecise probability.
\end{itemize}
We then compare  CDT, CMA and  COMP-AODE* via the Friedman test; this is the approach recommended by \citep{demsar06jmlr} for comparing multiple classifiers on multiple data sets. First, the procedure ranks on each data set the classifiers according to the utility they generate; then, it tests the null hypothesis of all classifiers having the same average rank across the data sets. If the null hypothesis is rejected, a post-hoc test is adopted to  identify
the significant differences among  classifiers.
Adopting a 95\% confidence,  no significant difference is detected among classifiers; the result is the same under both utilities.
 However, under both utilities COMP-AODE* has the best average rank, as shown in Figure \ref{fig:compcred}. Lowering the confidence to 90\%, two significant differences are found:
a) COMP-AODE* produces significantly higher utility than CMA under $u_{65}$ and b) COMP-AODE* produces significantly higher utility than CDT under $u_{80}$.
These results, though not completely conclusive, suggest that COMP-AODE* compares favorably to  previous credal classifiers.

\subsection{Some comments on credal classification vs reject option}
Determinate classifiers can be equipped with a \textit{reject option} \citep{herbei2006classification}, thus
 refusing to classify an instance if the posterior probability of the most probable class is less than a threshold.
For the sake of simplicity we consider a case with two classes only; to formally introduce the reject option, it  is necessary setting a cost $d$ ($0<d<1/2$), which is incurred when rejecting an instance. A cost $0$, $1$, $d$ is therefore incurred when respectively correctly classifying, wrongly classifying and rejecting an instance. Under 0-1 loss, the \textit{expected} cost for classifying an instance corresponds to the probability of misclassification; it is thus $1-p^*$, where  $p^*$ denotes the posterior probability of the most probable class. The optimal behavior is thus to reject the classification whenever the expected classification cost is higher than the rejection cost, namely when
$(1-p^*)>d$; this is equivalent to rejecting the instance whenever $p^*<1-d$, where ($1-d$) constitutes the \textit{rejection threshold}.

The behavior induced by the reject option is quite different from that of a credal classifier, as we show in the following example.
On  an a very large data set the posterior probability of the classes is little sensitive on the choice of the 
prior, because of the wide amount of data available for learning; in this condition, instance are rarely prior-dependent and therefore a credal classifier will mostly return a single class.
On the other hand,  the determinate classifier with reject option (RO in the following) rejects all the instances for which $p^*<1-d$; if $d$ is small, there can be even a high number of rejected instances.
The difference between these behaviors is due to the credal classifier being unaware of the cost $d$ associated with rejecting an instance, which is instead driving 
the behavior of RO. To rigorously  compare RO against a credal classifier, it is thus necessary making the credal classifier aware of the cost $d$.
Recalling that  the credal classifier already returns both classes on the instances which are prior dependent, this will change the behavior of the credal classifier only on the instances which are \textit{not} prior-dependent.
In particular, the credal classifier should reject all the instances for which $\underline{p}^*<1-d$, where $\underline{p}^*$ is the \textit{lower} probability of the most probable class; the instances rejected by means of  this criterion will be thus a superset of those rejected by RO. Therefore, the credal classifier will reject the instances which are prior-dependent \textit{and} those for which $\underline{p}^*<1-d$.
Eventually, the cost generated by the credal classifier should be compared with those generated by the RO.
In the case with more than 2 classes the analysis might become slightly more complicated than what discussed here; however,
we leave the analysis of credal classifiers with reject option as a topic for future research. Note also that this kind of experiment will require the computation of upper and lower posterior probability of the classes, which is not always trivial with credal classifiers.

\section{Conclusions}
Applying Bayesian Model Averaging over SPODEs  actually worsens the classification performance compared to the standard AODE. Instead the COMP-AODE classifier proposed here, which applies the compression-based approach over SPODEs, obtains overall slightly better classification performance than AODE; our results thus broadens the scope of \citep{boulle2007compression}, in which the compression-based approach was applied over an ensemble of naive Bayes classifiers.  The two credal classifiers BMA-AODE* and COMP-AODE* extend respectively BMA-AODE and COMP-AODE to imprecise probability,  replacing the uniform  prior over the SPODEs by a \textit{credal set}; both credal classifiers automatically identify the prior-dependent instances, and cope reliably with them by returning a small-sized but highly accurate set of classes. On the prior-dependent instances both BMA-AODE and COMP-AODE undergo a severe drop of accuracy.  Both BMA-AODE* and COMP-AODE*  provide overall higher performance than their determinate counterparts as measured by the utility-based measures, which to our knowledge constitute the state of the art for comparing determinate and credal classifiers. According to the same metrics, COMP-AODE* shows better performance than previous credal classifiers.

\section*{Acknowledgements}
The research in this paper has been partially supported by the Swiss NSF grants no. 200020-132252 and  by the Hasler foundation grant n.~10030.

% \bibliographystyle{elsarticle-harv}
% \bibliography{biblio}

\appendix
\newpage
\section{Mapping linear-fractional programs to linear programs by the Charnes-Cooper transformation}\label{app:cc}
In this appendix, we adapt the classical Charnes-Cooper transformation to the particular linear-fractional program to be solved
to test dominance for the BMA-AODE* as described in Section \ref{sec:cma}. Let us write the optimization variables as $x_j:=P(s_j)$ (with $j=1,\ldots,k$) and the coefficients as:
\begin{equation}
\left[ \begin{array}{c} \gamma_i \\ \delta_i \end{array} \right]
:= \left[ \begin{array}{c} P(c'|\mathbf{\tilde{a}},m_j)\\ P(c''|\mathbf{\tilde{a}},m_j)\end{array} \right]
\cdot L_j.
\end{equation}
The objective function rewrites therefore as:
\begin{equation}\label{eq:lf}
\frac{\sum_{j=1}^k \gamma_j x_j}{\sum_{j=1}^k \delta_j x_j}.
\end{equation}
with $j=1,\ldots,k$. Let us indeed change the variables as follows:
\begin{equation}
y_j  := \frac{x_j}{\sum_j \delta_j x_j},
\end{equation}
and introduce the auxiliary variable
\begin{equation}
t  := \frac{1}{\sum_j \delta_j x_j}.
\end{equation}
After this, non-linear, transformation, the objective function takes a linear form:
\begin{equation}
\sum_j \gamma_j y_j,
\end{equation}
while each linear constraint $x_j \geq \epsilon$, rewrites as $y_j \geq \epsilon t$, thus being still linear. Similarly, the normalization rewrites as:
$$\sum_j y_j = t.$$
We have therefore mapped the original problem into a standard linear program and the solutions of the two problems are known to coincide \cite[Chap. 3]{bajalinov2003linear}. Note that the transformation only increases by one the number of constraints.
\newpage
\section{Data sets list}
\begin{table}[!h]
\caption{\label{tab:dsets}List of the 40 data sets used for experiments.}
\begin{centering}
\begin{tabular}{cccccccccc}\toprule
dataset & $n$ & $k$ & classes &  &  & dataset & $n$ & $k$ & classes\\ \midrule
& & & &  &  & & & & \\
labor & 57 & 11 & 2 &  &  & ecoli & 336 & 6 & 8\tabularnewline
white\_clover & 63 & 6 & 4 &  &  & liver\_disorders & 345 & 1 & 2\tabularnewline
postoperative & 90 & 8 & 3 &  &  & ionosphere & 351 & 33 & 2\tabularnewline
zoo & 101 & 16 & 7 &  &  & monks3 & 554 & 6 & 2\tabularnewline
lymph & 148 & 18 & 4 &  &  & monks1 & 556 & 6 & 2\tabularnewline
iris & 150 & 4 & 3 &  &  & monks2 & 601 & 6 & 2\tabularnewline
tae & 151 & 2 & 3 &  &  & credit\_a & 690 & 15 & 2\tabularnewline
grub\_damage & 155 & 6 & 4 &  &  & breast\_w & 699 & 9 & 2\tabularnewline
hepatitis & 155 & 16 & 2 &  &  & diabetes & 768 & 6 & 2\tabularnewline
hayes\_roth & 160 & 3 & 3 &  &  & anneal & 898 & 31 & 6\tabularnewline
wine & 178 & 13 & 3 &  &  & credit\_g & 1000 & 15 & 2\tabularnewline
sonar & 208 & 21 & 2 &  &  & cmc & 1473 & 9 & 3\tabularnewline
glass & 214 & 7 & 7 &  &  & yeast & 1484 & 7 & 10\tabularnewline
heart\_h & 294 & 9 & 2 &  &  & segment & 2310 & 18 & 7\tabularnewline
heart\_c & 303 & 11 & 2 &  &  & kr\_vs\_kp & 3196 & 36 & 2\tabularnewline
haberman & 306 & 2 & 2 &  &  & hypothyroid & 3772 & 25 & 4\tabularnewline
solarflare\_C & 323 & 10 & 3 &  &  & waveform & 5000 & 19 & 3\tabularnewline
solarflare\_M & 323 & 10 & 4 &  &  & page\_blocks & 5473 & 10 & 5\tabularnewline
solarflare\_X & 323 & 10 & 2 &  &  & pendigits & 10992 & 16 & 10\tabularnewline
ecoli & 336 & 6 & 8 &  &  & nursery & 12960 & 8 & 5\\ \bottomrule
\end{tabular}
\par\end{centering}
\end{table}

\end{document}